\documentclass[nonacm,sigconf]{acmart}
\renewcommand\footnotetextcopyrightpermission[1]{} 

\AtBeginDocument{%
  \providecommand\BibTeX{{%
    \normalfont B\kern-0.5em{\scshape i\kern-0.25em b}\kern-0.8em\TeX}}}

\usepackage{multirow,tabularx}
\usepackage{makecell}

\usepackage{float}
\usepackage{subfig}

\usepackage[font=small]{caption}

\begin{document}

\title{On Adversarial Examples for Biomedical NLP Tasks}


 \author{Vladimir Araujo \and Andr\'es Carvallo \and Carlos Aspillaga \and Denis Parra}
\affiliation{Pontificia Universidad Cat\'olica de Chile}



\begin{abstract}
The success of pre-trained word embeddings has motivated its use in tasks in the biomedical domain. The BERT language model has shown remarkable results on standard performance metrics in tasks such as Named Entity Recognition (NER) and Semantic Textual Similarity (STS), which has brought significant progress in the field of NLP. However, it is unclear whether these systems work seemingly well in critical domains, such as legal or medical. For that reason, in this work, we propose an adversarial evaluation scheme on two well-known datasets for medical NER and STS. We propose two types of attacks inspired by natural spelling errors and typos made by humans. We also propose another type of attack that uses synonyms of medical terms. Under these adversarial settings, the accuracy of the models drops significantly, and we quantify the extent of this performance loss. We also show that we can significantly improve the robustness of the models by training them with adversarial examples. We hope our work will motivate the use of adversarial examples to evaluate and develop models with increased robustness for medical tasks.  

\end{abstract}




\maketitle

\section{Introduction}

Biomedical Natural Language Processing, or also known as BioNLP, are computational tools and methods for processing textual information and generally applied to tasks such as information retrieval, document classification, or literature-based discovery. Some applications of these techniques include gen-disease technology \cite{pletscher2015diseases}, the development of new drugs \cite{tari2010discovering}, or automatic screening of biomedical documents  \cite{carvallocomparing}.

With the exponential growth of digital biomedical literature, the application of natural language processing for decision-making is increasingly necessary. In order to encourage the development of this area, public datasets and challenges have been shared with the community to solve these tasks, such as BioSSES \cite{souganciouglu2017biosses}, HOC \cite{hoc2000}, ChemProt \cite{chemprot2016} and BC5CDR \cite{wei2015overview}. 

At the same time, general-purpose neural language models have recently shown significant progress with the introduction of models such as ELMo \cite{peters2018deep} and BERT \cite{devlin2018bert}.
These models have obtained remarkable results in tasks like Named Entity Recognition, Sentence Similarity and Multi-label Classification. A natural choice, therefore, is to apply these models to biomedical NLP.
Despite the impressive results achieved by the NLP models mentioned above, previous work has tested these models robustness by using adversarial attacks, showing that they are fragile under certain test conditions. This procedure consists of applying intentional perturbations to the input sentences and test whether they confuse a model into making wrong predictions or not. This methodology has shown that models are still weak and have limited ability to generalize or to understand the tasks they are dealing with \cite{belinkov2018synthetic,naik18coling,jin2019bert,aspillaga2020stress}. Moreover, in the domain of automated medical decision making, the robustness of the models being used is even more critical as they may impact on the well-being of patients \cite{hengstler2016applied}.

In this work, we focus on the BERT language model to carry out an adversarial evaluation on two BioMedical text mining tasks. On the one hand, Named Entity Recognition (NER) which consists in finding the relation between medical entities in text. On the other hand, semantic textual similarity (STS), that consists in deciding if two biomedical sentences are semantically related.  

The evaluation of the NER task was performed on the BC5CDR-disease and BC5CDR-chemical datasets \cite{wei2015overview}. In the case of STS, we used the BioSSES dataset \cite{souganciouglu2017biosses}. In this paper, we contribute by:
\begin{itemize}
    \item Testing the strength of pre-trained BERT for medical tasks under the scheme of adversarial attacks. 
    \item Demonstrating that the use of the proposed adversarial examples during training can increase the robustness of the model.
\end{itemize}

\begin{table*}[]
\caption{Adversarial Evaluation Sentence Examples}
\vspace{-3mm}
\small
\begin{tabular}{ll}
\toprule
\textbf{Original}            & Two mothers with heart valve prosthesis were treated with warfarin during pregnancy.           \\ \hline
\textbf{Swap Noise}          & Two mothers with \textcolor{blue}{herat vavle protshesis} were \textcolor{blue}{terated} with \textcolor{blue}{warafrin} during \textcolor{blue}{preganncy}.           \\ \hline
\textbf{Keyboard Typo Noise} & Two mothers with \textcolor{blue}{hea5t valce prosth3sis} were \textcolor{blue}{trezted} with \textcolor{blue}{warfsrin} during \textcolor{blue}{pregnahcy}.           \\ \hline
\textbf{Synonymy}            & Two mothers with heart valve prosthesis were treated with \textcolor{blue}{potassium warfarin} during pregnancy. \\ 
\bottomrule
\end{tabular}
\vspace{-4mm}
\label{table0}
\end{table*}

\section{Adversarial Examples}
Previous work on adversarial examples has demonstrated how dangerous it can be to use machine learning systems in real-world applications \cite{42503,goodfellow2014explaining}. This evaluation strategy showed that slight disturbances in the inputs could cause severe failures in deep neural networks. Early work proposed methodologies to generate adverse examples in order to evaluate computer vision models \cite{Akhtar_2018}. As a result, it was demonstrated the weakness of the systems and proposing a defense mechanism through adversarial training \cite{goodfellow2014explaining}.

Perturbation methods developed for images cannot be directly applied to texts in most cases. Because of this, recent work has explored new adversarial strategies specially designed for NLP tasks \cite{zhang2019adversarial}. The evaluation of NLP models has been carried out mainly in trained models for a single task \cite{belinkov2018synthetic,naik18coling}. More recently, due to the success of pre-trained models based on recurrent and transformer-based networks, adversarial attacks have been applied to several NLP benchmarks \cite{jin2019bert,aspillaga2020stress}. 

Models used in the medical tasks have noticed particular interest because an erroneous prediction could be very harmful to a patient. Most of the proposed adversarial attacks for the medical tasks are related to image analysis \cite{Finlayson2019,ma2019understanding}. Despite the existence of deployed systems in real-world clinical settings, researchers have shown that even the state of the art models in medical computer vision is vulnerable to adversarial attacks. For this reason, the use of this type of methodology has been encouraged for the evaluation and improvement of models to avoid critical errors in these scenarios.

There is a growing availability of resources for biomedical NLP tasks, which are mainly resolved through the use of pre-trained models \cite{peng2019transfer}. However, as far as we know, no work evaluates such models under adversarial attacks. For that reason, in this paper, we focus on assessing the state of the art BioNLP models robustness and study how to strengthen them through adversarial training.

\section{Biomedicine Text Mining Tasks}
We chose two tasks from the recently introduced BLUE benchmark \cite{peng2019transfer}. Not all medical resources are made public, so we decided to focus only on the tasks that have publicly accessible datasets (BCD5CDR-Chemical, BC5CDR-Disease, and BioSSES)\footnote{https://biocreative.bioinformatics.udel.edu/tasks/biocreative-v/track-3-cdr/}.

\textbf{Named Entity Recognition}. NER is a subtask of information extraction that seeks to locate named entities in text and classify them into pre-defined medical categories, such as protein, cell type, chemical, disease, and so forth.

For this task, we use the BC5CDR dataset \cite{wei2015overview}, which consists of 1500 PubMed articles with 4409 annotated chemicals and 5818 diseases. They were selected from the CTD-Pfizer corpus that was used in the BioCreative V chemical-disease relation task.

Given each word in a sentence, the goal of the model is to predict its labels, following the IOB (Inside-outside-beginning) format. We use the two variations of the dataset, one related to diseases and the other to chemical components.

\textbf{Semantic Textual Similarity}. STS measures the degree of equivalence in the underlying semantics of paired snippets of text. The task is to predict the similarity score between two medical sentences. 

We use the BioSSES\footnote{https://tabilab.cmpe.boun.edu.tr/BIOSSES/} dataset \cite{souganciouglu2017biosses}, which is a corpus of sentence pairs selected from the Biomedical Summarization Track. The objective of BioSSES is to compute the similarity of biomedical sentences by utilizing WordNet as a general domain ontology and Unified Medical Language System (UMLS) as biomedical domain-specific ontology. The main task is to estimate a similarity score between sentence pairs that have been manually validated by physicians.

\section{Biomedical LM \& Adversarial Evaluation}
\paragraph{Biomedical Language Models} For this test, we focus on a pre-trained BERT model because it is well known for its outstanding performance in several domains of NLP. Recent BioNLP community efforts allowed the specialization of this type of models by training on a large medical corpus \cite{peng2019transfer}. In order to use this model for the selected tasks, we follow the fine-tuning procedure and evaluation metrics proposed on the BLUE benchmark \cite{peng2019transfer}.
We compare the results of BERT pre-trained on PubMed abstracts with the same model but pre-trained on PubMed and MIMIC-III datasets.

\subsection{Adversarial Evaluation}
\label{adv_eval}
For this evaluation, we propose a black-box attack methodology, which does not require the inner details of the model to generate adverse examples \cite{zhang2019adversarial}. Specifically, we focus on making disturbances in the input data, also known as edit adversaries, that could cause the models to fall into erroneous predictions. The following subsections describe each of the adversarial sets, and their construction\footnote{Adversarial dataset will be available after notification}. Also, we show examples of the adversarial attacks in Table~\ref{table0}.

\paragraph{\textbf{Noise Adversaries}}
This evaluation aims to test the robustness of models to spelling errors. Misspelling is not usual in medical resources; however, it is possible to come across it \cite{Pinto,Lai2015}. Motivated by the above and inspired by \cite{belinkov2018synthetic}, we constructed adversarial examples that try to emulate spelling errors committed by human beings. 
We use spaCy models \cite{Neumann2019ScispaCyFA} for processing texts of the datasets to retrieve the medical terms of each sentence. Then, each term is replaced by a noisy word. These edit adversaries consist of two types of alterations: (i) \textbf{Swap Noise}: For each word, one random pair of consecutive characters is swapped, (ii) \textbf{Keyboard Typo Noise}: For each word, one character is replaced by an adjacent character in traditional English keyboards.

\paragraph{\textbf{Synonymy Adversaries}}
These examples test if a model can understand synonymy relations. Replacing a medical term with an equivalent synonym is challenging. For that reason, we focus only on words of chemicals and diseases.

We use spaCy models to identify chemical and disease entities of sentences. Then we use PyMedTermino \cite{Lamy}, which uses the biomedical vocabulary of UMLS, to find the most similar or related words (synonyms) to the retrieved words. Finally, we replace the synonym found depending on whether it is a disease or chemical.

\begin{table}[]
\caption{Adversarial evaluation in the NER task.}
\vspace{-3mm}
\setlength{\tabcolsep}{0.45em} 
{\renewcommand{\arraystretch}{0.95}
\label{table-ner}
\centering
{
\small
\begin{tabular}{ccccccccc} \toprule
    {Train Set} & {Model} & {Test Set} & {Precision} & {Recall} & {F1}    \\ \hline
    
    \multirow{8}{*}{\makecell{BC5CDR- \\ Chemical}} & {} & {Original}  & \textbf{.894} & \textbf{.893} & .\textbf{894}   \\
    {} & {BERT} & {Synonym} & .744 & .739 & .741    \\ 
    {} & {P+M} & {Keyboard} & .619 & .541 & .578   \\ 
    {} & {} & {Swap} & .764 & .698 & .728   \\ 
   
    \cline{2-6}
    
    \multirow{8}{*}{} & {} & {Original}  & \textbf{.895} & \textbf{.908} & \textbf{.901}  \\
    {} & {BERT} & {Synonym} & .730 & .748 & .739 \\ 
    {} & {P} & {Keyboard} & .734 & .683 & .708    \\ 
    {} & {} & {Swap} & .609 & .559 & .583   \\ 
    
   
    \hline
    
    \multirow{8}{*}{\makecell{BC5CDR- \\ Disease}} & {} & {Original}  & \textbf{.828} & \textbf{.829} &  \textbf{.828}  \\
    {} & {BERT} & {Synonym} & .322  & .371 & .345 \\ 
    {} & {P+M} & {Keyboard} & .528 & .273 & .360   \\ 
    {} & {} & {Swap} & .636 & .369 & .467  \\ 
    \cline{2-6}
    
    \multirow{8}{*}{} & {} & {Original}  & \textbf{.832} & \textbf{.844} & \textbf{.838}   \\
    {} & {BERT} & {Synonym} & .337 & .390 & .362 \\ 
    {} & {P} & {Keyboard} & .543 & .278 & .368   \\ 
    {} & {} & {Swap} & .636 & .337 & .441  \\ 
    
    
   
   \bottomrule
    \end{tabular}}}
\vspace{-5mm}    
\end{table}

\begin{figure}[t]
  \caption{Adversarial evaluation in the NER task. }
  \vspace{-3mm}
    \centering
    \subfloat[BC5CDR-Chemical]{
    \includegraphics[scale=0.35,trim={0mm 7mm 0mm 7mm},clip]{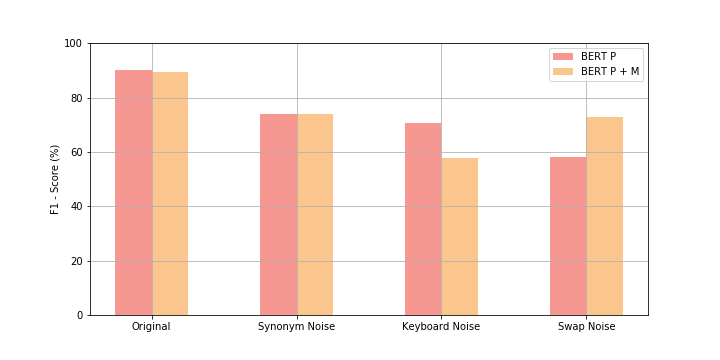}}
    \qquad \vspace{-5mm}
    \subfloat[BC5CDR-Disease]{{\includegraphics[scale=0.35,,trim={0mm 7mm 0mm 15mm},clip]{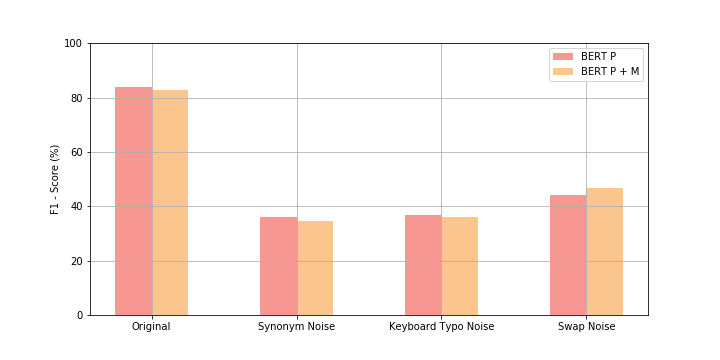}}}%
    \label{fig_stress_NER}
\vspace{0mm}
\end{figure}

\section{Experiments}
\paragraph{\textbf{Experimental Setup}} We use two BERT-base models pre-trained on PubMed (P) or PubMed and MIMIC-III (P+M) datasets \cite{peng2019transfer}. We fine-tune the models with the original train sets of each task and evaluate them with the original set test and the adversarial sets.

In terms of metrics, since NER is a classification task, we measure the F1 score. However, for STS experiments, we evaluate similarity by using Pearson correlation coefficients.

\paragraph{\textbf{Results on Adversarial Evaluation for the NER Task}} Table~\ref{table-ner} shows the classification results of the BC5CDR task on our adversarial examples and the original test set. We see that the performance of BERT drops across all adversarial attacks. However, the task of recognizing the disease was the most affected.

In the case of the chemical recognition task (Figure~\ref{fig_stress_NER}a), both BERT versions show a drop of approximately 20\% of the F1 score. In general, the drop in the model is slight, which could be because the number of annotated chemicals is smaller than the number of annotated diseases.

In contrast, the F1 score of the disease recognition model (Figure~\ref{fig_stress_NER}b) falls dramatically, below 50\% of the original score. As we mentioned above, the disease set contains more annotations. For this reason, its adversarial dataset becomes harder as noise adversaries are applied to all the annotations. If there are more diseases present in the dataset, the model will be exposed to more adversarial attacks.

Comparing the performance of BERT (P+M) with BERT (P), on the original test set they achieve similar F1 scores. Then when adding keyboard noise, BERT P is less affected than BERT P+M. In the case of synonym noise, it affects both models in similar proportions. Finally, when adding swap noise, the effect is contrary, BERT P+M is less affected than BERT P. 
This pattern is observed in both Chemical and Disease NER datasets.

\paragraph{\textbf{Results on Adversarial Evaluation for the STS Task}} Table~\ref{table-sts} shows the results of the STS task tested in the BioSSES dataset. We see that the Pearson coefficient of the model drops about 23\% (P+M) and 8\% (P) under adversarial noise attacks.
However, the results of synonymy adversaries show a different response for both models. This task comprises two sentences, but we modify only one of them. We hypothesize that words replaced by their synonyms in one sentence tend to be the same or more similar to the other sentence. It would explain why the Pearson coefficient with synonymy adversaries is higher than with the original test set.

\begin{table}[!t]
\caption{Stress Semantic Textual Similarity Task tests results using adversarial test set.}
\vspace{-3mm}
\setlength{\tabcolsep}{0.45em} 
{\renewcommand{\arraystretch}{0.95}
\label{table-sts}
\centering
{
\small
\begin{tabular}{ccccccccc} \toprule
    {Train Set} & {Model} & {Test Set}  & {Pearson} & {Spearman}    \\ \hline
    
    \multirow{4}{*}{BioSSES} & {} & {Original}    & .832 & .744   \\
    {} & {BERT} & {Synonym} &  \textbf{.844} & \textbf{.774} \\ 
    {} & {P+M} & {Keyboard}  & .656 & .656   \\ 
    {} & {} & {Swap}  & .622 & .685   \\ 
   
   \hline
   
   \multirow{4}{*}{BioSSES} & {} & {Original}   & .829 & \textbf{.813}   \\
    {} & {BERT} & {Synonym} & \textbf{.869} & .666 \\ 
    {} & {P} & {Keyboard}  & .759 & .607   \\ 
    {} & {} & {Swap}  & .765 & .774  \\ 
   
   \bottomrule
    \end{tabular}}}
\vspace{-5mm}    
\end{table}

\paragraph{\textbf{Adversarial Training Results}} Training with adversarial examples is a methodology used in previous works \cite{belinkov2018synthetic,jia-liang-2017-adversarial} to create robustness in neural language models. It ensures that the model is exposed to samples outside the training distribution and provides a form of regularization \cite{belinkov2018synthetic}.

For both tasks, we first fine-tune the model with the original training set plus an adversarial version of the same set. Then we carry out the same procedure explained in Section~\ref{adv_eval}, to measure how the models perform in the different test sets.

On the one hand, Table~\ref{table-ner-che-train} and Table~\ref{table-ner-dis-train} show the results for NER of training with adversaries and testing with the original set compared with their respective adversaries.

On the other hand, Table~\ref{table-sts-train} we train for the STS task with adversaries and testing with different test sets.

In both cases, we see that training with adversarial examples significantly improves robustness of the models to adversarial attacks, without significant impact on the original non-adversarial task.

\begin{table}[!t]
\caption{Adversarial evaluation in the BC5CDR-Chemical NER task after adversarial training.}
\vspace{-3mm}
\setlength{\tabcolsep}{0.45em} 
{\renewcommand{\arraystretch}{0.95}
\label{table-ner-che-train}
\centering
{
\small
\begin{tabular}{cccccccccc} \toprule
    {Model} & {Train} & {Test} & {precision} & {recall} & {f1}    \\ 
    \hline
    \multirow{6}{*}{\makecell{BERT \\ P+M}} & \multirow{2}{*}{Synonym} & {Original}  & .888 & .886  & .887    \\
    {} & {} & {Synonym} & .863 & .898  & .880   \\ 
    \cline{2-6}
    {} & \multirow{2}{*}{Keyboard} & {Original}  & .892 & .890 & .891   \\
    {} & {} & {Keyboard} & .834 & .810 & .822\\ 
    \cline{2-6}
    {} & \multirow{2}{*}{Swap} & {Original}  & .886 & .889  & .888   \\
    {} & {} & {Swap} & .668 & .610 & .638 \\ 
    
    \hline
    
    \multirow{6}{*}{\makecell{BERT \\ P}} & \multirow{2}{*}{Synonym} & {Original}  & \textbf{.899} & .901  & \textbf{.900}   \\
    {} & {} & {Synonym} & .872 & \textbf{.908}  & .890 \\ 
    \cline{2-6}
    {} & \multirow{2}{*}{Keyboard} & {Original}  & .889 & .906 & .898\\
    {} & {} & {Keyboard} & .850 & .792 & .820 \\ 
    \cline{2-6}
    {} & \multirow{2}{*}{Swap} & {Original}  & .895 & .902 & .898\\
    {} & {} & {Swap} & .684 & .630 & .656 \\ 
    
    
    \bottomrule
\end{tabular}}}
\vspace{-3mm}
\end{table}

\begin{table}[!ht]
\caption{Adversarial evaluation in the BC5CDR-Disease NER task after adversarial training.}
\vspace{-3mm} 
\setlength{\tabcolsep}{0.45em} 
{\renewcommand{\arraystretch}{0.95}
\label{table-ner-dis-train}
\centering
{
\small
\begin{tabular}{ccccccccc} \toprule
    {Model} & {Train} & {Test} & {precision} & {recall} & {f1}    \\ 
    \hline
    \multirow{6}{*}{\makecell{BERT \\ P+M}} & \multirow{2}{*}{Synonym} & {Original}  & .805 & .806 & .805\\
    {} & {} & {Synonym} & .786 & .832 & .808   \\ 
    \cline{2-6}
    {} & \multirow{2}{*}{Keyboard} & {Original}  & .830 & .839 & .835   \\
    {} & {} & {Keyboard} & .711 & .698 & .704\\ 
    \cline{2-6}
    {} & \multirow{2}{*}{Swap} & {Original}  & .826 & .831  & .828   \\
    {} & {} & {Swap} & .745 & .736 & .741 \\ 
    \hline
    
    \multirow{6}{*}{\makecell{BERT \\ P}} & \multirow{2}{*}{Synonym} & {Original}  & .813 & .824  & .818   \\
    {} & {} & {Synonym} & .788 & .841  & .814 \\ 
    \cline{2-6}
    {} & \multirow{2}{*}{Keyboard} & {Original}  & \textbf{.839} & \textbf{.848} & \textbf{.844}\\
    {} & {} & {Keyboard} & .723 & .712 & .717 \\ 
    \cline{2-6}
    {} & \multirow{2}{*}{Swap} & {Original}  & .836 & .847 & .841    \\
    {} & {} & {Swap} & .773 & .746 & .759 \\ 
    
    
    \bottomrule
\end{tabular}}}
\vspace{-3mm}
\end{table}

\begin{table}[!h]
\caption{Adversarial evaluation in the STS task after adversarial training.} 
\vspace{-3mm}
\setlength{\tabcolsep}{0.45em} 
{\renewcommand{\arraystretch}{0.95}
\label{table-sts-train}
\centering
{
\small
\begin{tabular}{ccccccccc} \toprule
    {Model} & {Train} & {Test} & {Pearson} & {Spearman}    \\ 
    \hline
    \multirow{6}{*}{\makecell{BERT \\ P+M}} & \multirow{2}{*}{Synonym} & {Original}  & .723 & .607\\
    {} & {} & {Synonym} & .742 & .557\\ 
    \cline{2-6}
    {} & \multirow{2}{*}{Keyboard} & {Original}  & .741 & .626  \\
    {} & {} & {Keyboard} & .676 & .538 \\ 
    \cline{2-6}
    {} & \multirow{2}{*}{Swap} & {Original}  & .615 & .557   \\
    {} & {} & {Swap} & .528 & .508 \\ 
    \hline
    
    \multirow{6}{*}{\makecell{BERT \\ P}} & \multirow{2}{*}{Synonym} & {Original}  & .790  & .518     \\
    {} & {} & {Synonym} & .796 & .705   \\ 
    \cline{2-6}
    {} & \multirow{2}{*}{Keyboard} & {Original}  & \textbf{.827} & .715     \\
    {} & {} & {Keyboard} & .789 & .607 \\ 
    \cline{2-6}
    {} & \multirow{2}{*}{Swap} & {Original}  & .566 & .734      \\
    {} & {} & {Swap} & .775 & \textbf{.774} &  \\ 
    
    
    \bottomrule
\end{tabular}}}
\vspace{-3mm}
\end{table}

\section{Conclusions} 
By observing previous results obtained by state-of-the-art language models, the NER and STS tasks seemed close to being solved. However, when applying adversaries, we realize the need for adversarial evaluation and training to ensure the strength of the models.

We noted that a relevant factor is the content used to train the BERT language model. Theoretically, a more extensive vocabulary would give the model a higher capability to generalize for new cases. However, we observed the opposite: in most of the cases, training only with PubMed allowed the model to be more prepared to adversarial attacks than when pre-trained with PubMed and MIMIC-III.

Unexpected behaviors were observed when replacing medical terms with synonyms in the STS task. Adversarial synonym tests surpassed the original results. We hypothesize that it might be because the adversaries introduced by us in some cases brought together terms that in the original dataset were further apart.

For future work, we plan to explore other medicine-related tasks, such as document-screening or multi-label classification. 


\bibliographystyle{ACM-Reference-Format}
\bibliography{sample-base}


\end{document}